\catcode`\@=11					% To make protected \def's

%************************************************************
%*
%*		Font set-up
%*
%************************************************************

%************** 5-point fonts *******************************

\font\fiverm=cmr5				% roman
\font\fivemi=cmmi5				% math italic
\font\fivesy=cmsy5				% math symbols
\font\fivebf=cmbx5				% bold face

\skewchar\fivemi='177
\skewchar\fivesy='60

%************** 6-point fonts *******************************

\font\sixrm=cmr6				% roman
\font\sixi=cmmi6				% math italic
\font\sixsy=cmsy6				% math symbols
\font\sixbf=cmbx6				% bold face

\skewchar\sixi='177
\skewchar\sixsy='60

%************** 7-point fonts *******************************

\font\sevenrm=cmr7				% roman
\font\seveni=cmmi7				% math italic
\font\sevensy=cmsy7				% math symbols
\font\sevenit=cmti7				% italic
\font\sevenbf=cmbx7				% bold face

\skewchar\seveni='177
\skewchar\sevensy='60

%************** 8-point fonts *******************************

\font\eightrm=cmr8				% roman
\font\eighti=cmmi8				% math italic
\font\eightsy=cmsy8				% math symbols
\font\eightit=cmti8				% italic
				% slanted
\font\eightbf=cmbx8				% bold face
				% typewriter
				% sans serif

\skewchar\eighti='177
\skewchar\eightsy='60

%************** 9-point fonts *******************************

\font\ninei=cmmi9
\font\ninesy=cmsy9

\skewchar\ninei='177
\skewchar\ninesy='60

%************** 10-point fonts ******************************

\font\tenrm=cmr10				% roman
\font\teni=cmmi10				% math italic
\font\tensy=cmsy10				% math symbols
\font\tenex=cmex10				% math extension
\font\tenit=cmti10				% italic
\font\tensl=cmsl10				% slanted
\font\tenbf=cmbx10				% bold face
\font\tentt=cmtt10				% typewriter
\font\tenss=cmss10				% sans serif
\font\tensc=cmcsc10				% small caps
\font\tenbi=cmmib10				% bold math

\skewchar\teni='177
\skewchar\tenbi='177
\skewchar\tensy='60

\def\tenpoint{\ifmmode\err@badsizechange\else
	\textfont0=\tenrm \scriptfont0=\sevenrm \scriptscriptfont0=\fiverm
	\textfont1=\teni  \scriptfont1=\seveni  \scriptscriptfont1=\fivemi
	\textfont2=\tensy \scriptfont2=\sevensy \scriptscriptfont2=\fivesy
	\textfont3=\tenex \scriptfont3=\tenex   \scriptscriptfont3=\tenex
	\textfont4=\tenit \scriptfont4=\sevenit \scriptscriptfont4=\sevenit
	\textfont5=\tensl
	\textfont6=\tenbf \scriptfont6=\sevenbf \scriptscriptfont6=\fivebf
	\textfont7=\tentt
	\textfont8=\tenbi \scriptfont8=\seveni  \scriptscriptfont8=\fivemi
	\def\rm{\tenrm\fam=0 }%
	\def\it{\tenit\fam=4 }%
	\def\sl{\tensl\fam=5 }%
	\def\bf{\tenbf\fam=6 }%
	\def\tt{\tentt\fam=7 }%
	\def\ss{\tenss}%
	\def\sc{\tensc}%
	\def\bmit{\fam=8 }%
	\rm\setparameters\setbaselines\fi}

%************** 12-point fonts ******************************

\font\twelverm=cmr12				% roman
\font\twelvei=cmmi12				% math italic
\font\twelvesy=cmsy10	scaled\magstep1		% math symbols
\font\twelveex=cmex10	scaled\magstep1		% math extension
\font\twelveit=cmti12				% italic
\font\twelvesl=cmsl12				% slanted
\font\twelvebf=cmbx12				% bold face
\font\twelvett=cmtt12				% typewriter
\font\twelvess=cmss12				% sans serif
\font\twelvesc=cmcsc10	scaled\magstep1		% small caps
\font\twelvebi=cmmib10	scaled\magstep1		% bold math

\skewchar\twelvei='177
\skewchar\twelvebi='177
\skewchar\twelvesy='60

\def\twelvepoint{\ifmmode\err@badsizechange\else
	\textfont0=\twelverm \scriptfont0=\eightrm \scriptscriptfont0=\sixrm
	\textfont1=\twelvei  \scriptfont1=\eighti  \scriptscriptfont1=\sixi
	\textfont2=\twelvesy \scriptfont2=\eightsy \scriptscriptfont2=\sixsy
	\textfont3=\twelveex \scriptfont3=\tenex   \scriptscriptfont3=\tenex
	\textfont4=\twelveit \scriptfont4=\eightit \scriptscriptfont4=\sevenit
	\textfont5=\twelvesl
	\textfont6=\twelvebf \scriptfont6=\eightbf \scriptscriptfont6=\sixbf
	\textfont7=\twelvett
	\textfont8=\twelvebi \scriptfont8=\eighti  \scriptscriptfont8=\sixi
	\def\rm{\twelverm\fam=0 }%
	\def\it{\twelveit\fam=4 }%
	\def\sl{\twelvesl\fam=5 }%
	\def\bf{\twelvebf\fam=6 }%
	\def\tt{\twelvett\fam=7 }%
	\def\ss{\twelvess}%
	\def\sc{\twelvesc}%
	\def\bmit{\fam=8 }%
	\rm\setparameters\setbaselines\fi}

%************** 14-point fonts ******************************

\font\fourteenrm=cmr12	scaled\magstep1		% roman
\font\fourteeni=cmmi12	scaled\magstep1		% math italic
\font\fourteensy=cmsy10	scaled\magstep2		% math symbols
\font\fourteenex=cmex10	scaled\magstep2		% math extension
\font\fourteenit=cmti12	scaled\magstep1		% italic
\font\fourteensl=cmsl12	scaled\magstep1		% slanted
\font\fourteenbf=cmbx12	scaled\magstep1		% bold face
\font\fourteentt=cmtt12	scaled\magstep1		% typewriter
\font\fourteenss=cmss12	scaled\magstep1		% sans serif
\font\fourteensc=cmcsc10 scaled\magstep2	% small caps
\font\fourteenbi=cmmib10 scaled\magstep2	% bold math

\skewchar\fourteeni='177
\skewchar\fourteenbi='177
\skewchar\fourteensy='60

\def\fourteenpoint{\ifmmode\err@badsizechange\else
	\textfont0=\fourteenrm \scriptfont0=\tenrm \scriptscriptfont0=\sevenrm
	\textfont1=\fourteeni  \scriptfont1=\teni  \scriptscriptfont1=\seveni
	\textfont2=\fourteensy \scriptfont2=\tensy \scriptscriptfont2=\sevensy
	\textfont3=\fourteenex \scriptfont3=\tenex \scriptscriptfont3=\tenex
	\textfont4=\fourteenit \scriptfont4=\tenit \scriptscriptfont4=\sevenit
	\textfont5=\fourteensl
	\textfont6=\fourteenbf \scriptfont6=\tenbf \scriptscriptfont6=\sevenbf
	\textfont7=\fourteentt
	\textfont8=\fourteenbi \scriptfont8=\tenbi \scriptscriptfont8=\seveni
	\def\rm{\fourteenrm\fam=0 }%
	\def\it{\fourteenit\fam=4 }%
	\def\sl{\fourteensl\fam=5 }%
	\def\bf{\fourteenbf\fam=6 }%
	\def\tt{\fourteentt\fam=7}%
	\def\ss{\fourteenss}%
	\def\sc{\fourteensc}%
	\def\bmit{\fam=8 }%
	\rm\setparameters\setbaselines\fi}

%************** Miscellaneous big fonts *********************

\font\seventeenrm=cmr10 scaled\magstep3		% roman
		% bold face

%************************************************************
%*
%*		Parameter initialization
%*
%************************************************************

\newdimen\rp@
\newcount\@basestretchnum
\newskip\@baseskip
\newskip\headskip
\newskip\footskip

% Routine to set page parameters

\def\setparameters{\rp@=.1em
	\headskip=24\rp@
	\footskip=\headskip
	\delimitershortfall=5\rp@
	\nulldelimiterspace=1.2\rp@
	\scriptspace=0.5\rp@
	\abovedisplayskip=10\rp@ plus3\rp@ minus5\rp@
	\belowdisplayskip=10\rp@ plus3\rp@ minus5\rp@
	\abovedisplayshortskip=5\rp@ plus2\rp@ minus4\rp@
	\belowdisplayshortskip=10\rp@ plus3\rp@ minus5\rp@
	\normallineskip=\rp@
	\lineskip=\normallineskip
	\normallineskiplimit=0pt
	\lineskiplimit=\normallineskiplimit
	\jot=3\rp@
	\setbox0=\hbox{\the\textfont3 B}\p@renwd=\wd0
	\skip\footins=12\rp@ plus3\rp@ minus3\rp@
	\skip\topins=0pt plus0pt minus0pt}

% Special routine to scale \baselineskip

\def\setbaselines{\maxdepth=4\rp@\baselinestretch=\@basestretchnum}

% The \baselinestretch command

\def\baselinestretch{\afterassignment\@basestretch\@basestretchnum}
\def\@basestretch{%
	\@baseskip=12\rp@ \divide\@baseskip by1000
	\normalbaselineskip=\@basestretchnum\@baseskip
	\baselineskip=\normalbaselineskip
	\bigskipamount=\the\baselineskip
		plus.25\baselineskip minus.25\baselineskip
	\medskipamount=.5\baselineskip
		plus.125\baselineskip minus.125\baselineskip
	\smallskipamount=.25\baselineskip
		plus.0625\baselineskip minus.0625\baselineskip
	\setbox\strutbox=\hbox{\vrule height.708\baselineskip
		depth.292\baselineskip width0pt }}

%************************************************************
%*
%*		Modifications to PLAIN.TEX
%*
%************************************************************

% Modifications to PLAIN routines to handle scaling of page parameters

\def\makeheadline{\vbox to0pt{\baselinestretch=1000
	\vskip-\headskip \vskip1.5pt
	\line{\vbox to\ht\strutbox{}\the\headline}\vss}\nointerlineskip}

\def\makefootline{\baselineskip=\footskip\line{\the\footline}}

\def\big#1{{\hbox{$\left#1\vbox to8.5\rp@ {}\right.\n@space$}}}
\def\Big#1{{\hbox{$\left#1\vbox to11.5\rp@ {}\right.\n@space$}}}
\def\bigg#1{{\hbox{$\left#1\vbox to14.5\rp@ {}\right.\n@space$}}}
\def\Bigg#1{{\hbox{$\left#1\vbox to17.5\rp@ {}\right.\n@space$}}}

% Modifications to PLAIN to handle bold math

\mathchardef\alpha="710B
\mathchardef\beta="710C
\mathchardef\gamma="710D
\mathchardef\delta="710E
\mathchardef\epsilon="710F
\mathchardef\zeta="7110
\mathchardef\eta="7111
\mathchardef\theta="7112
\mathchardef\iota="7113
\mathchardef\kappa="7114
\mathchardef\lambda="7115
\mathchardef\mu="7116
\mathchardef\nu="7117
\mathchardef\xi="7118
\mathchardef\pi="7119
\mathchardef\rho="711A
\mathchardef\sigma="711B
\mathchardef\tau="711C
\mathchardef\upsilon="711D
\mathchardef\phi="711E
\mathchardef\chi="711F
\mathchardef\psi="7120
\mathchardef\omega="7121
\mathchardef\varepsilon="7122
\mathchardef\vartheta="7123
\mathchardef\varpi="7124
\mathchardef\varrho="7125
\mathchardef\varsigma="7126
\mathchardef\varphi="7127
\mathchardef\imath="717B
\mathchardef\jmath="717C
\mathchardef\ell="7160
\mathchardef\wp="717D
\mathchardef\partial="7140
\mathchardef\flat="715B
\mathchardef\natural="715C
\mathchardef\sharp="715D

%************************************************************
%*
%*		Initialization
%*
%************************************************************

\def\err@badsizechange{%
	\immediate\write16{--> Size change not allowed in math mode, ignored}}

\baselinestretch=1000
\tenpoint

\catcode`\@=12					% Restore @ sign
% Routine to guarantee that this file is input only once
\catcode`\@=11
\expandafter\ifx\csname @iasmacros\endcsname\relax
	\global\let\@iasmacros=\par
\else	\immediate\write16{}
	\immediate\write16{Warning:}
	\immediate\write16{You have tried to input iasmacros more than once.}
	\immediate\write16{}
	\endinput
\fi
\catcode`\@=12

% Set up font size commands and \baselinestretch command
%\input iasfonts

% Some alternative font names
\def\rmb{\seventeenrm}

% Simple spacing commands
\def\singlespace{\baselineskip=\normalbaselineskip}
\def\halfspace{\baselineskip=1.5\normalbaselineskip}
\def\doublespace{\baselineskip=2\normalbaselineskip}

% Macros for references and abstracts

\def\AB{\bigskip\parindent=40pt
        \centerline{\bf ABSTRACT}\medskip\halfspace\narrower}
\def\AE{\bigskip\nonarrower\doublespace}
\def\nonarrower{\advance\leftskip by-\parindent
	\advance\rightskip by-\parindent}

% Useful commands

\def\boxit#1{\vbox{\hrule\hbox{\vrule\kern3pt
	\vbox{\kern3pt#1\kern3pt}\kern3pt\vrule}\hrule}}

% Special symbols
\def\hence{\leavevmode\hbox{\bf .\raise5.5pt\hbox{.}.} }

\def\dalemb#1#2{{\vbox{\hrule height.#2pt
	\hbox{\vrule width.#2pt height#1pt \kern#1pt \vrule width.#2pt}
	\hrule height.#2pt}}}
\def\gtorder{\mathrel{\raise.3ex\hbox{$>$}\mkern-14mu
             \lower0.6ex\hbox{$\sim$}}}
\def\ltorder{\mathrel{\raise.3ex\hbox{$<$}\mkern-14mu
             \lower0.6ex\hbox{$\sim$}}}

% For twoup output
\newdimen\fullhsize
\newbox\leftcolumn
\def\twoup{\hoffset=-.5in \voffset=-.25in
  \hsize=4.75in \fullhsize=10in \vsize=6.9in
  \def\fullline{\hbox to\fullhsize}
  \let\lr=L
  \output={\if L\lr
        \global\setbox\leftcolumn=\columnbox\global\let\lr=R \advancepageno
      \else \doubleformat \global\let\lr=L\fi
    \ifnum\outputpenalty>-20000 \else\dosupereject\fi}
  \def\doubleformat{\shipout\vbox{
    \fullline{\box\leftcolumn\hfil\columnbox}\advancepageno}}
  \def\columnbox{\leftline{\vbox{\makeheadline\pagebody\makefootline}}}
  \tolerance=1000 }

\twelvepoint
\doublespace

{{
%\rightline{IASSNS--HEP--95/89}
\rightline{~~~ October, 1998}
\bigskip\bigskip
\centerline{\rmb General Theory of Image Normalization}
\medskip
\centerline{\bf Stephen L. Adler
%{\singlespace { Research supported in part by the Department of Energy 
%under Grant No.
%DE--FG02--90ER40542.}}
}
\centerline{\bf Institute for Advanced Study}
\centerline{\bf Princeton, NJ 08540}
\medskip
\bigskip
%\leftline{\it Submitted to Computer Vision and Image Understanding}
\bigskip
\leftline{\it Send correspondence to:}
\medskip
{\singlespace\leftline{Stephen L. Adler}
\leftline{Institute for Advanced Study}
\leftline{Olden Lane, Princeton, NJ 08540}
\leftline{Phone 609--734--8051; FAX 609--924--8399; email adler@ias.edu}}
\bigskip\bigskip
}}
\vfill \eject
\bigskip
\AB
We give a systematic, abstract formulation of the image
normalization method as applied to a general group of image transformations, 
and then illustrate the abstract analysis by applying it to the hierarchy of  
viewing transformations of a planar object.
\AE
\bigskip\bigskip
\vfill\eject
\bigskip

\centerline{{\bf 1.~~Introduction and Brief Review of Viewing}}
\centerline{{\bf Transformations of a Planar Object}}

A central issue in pattern recognition is the efficient incorporation of
invariances with respect to geometric viewing transformations.    
We focus in this article on a particular method for handling invariances,
called ``image normalization'', which has the capability of extracting all of
the invariant features from an image using only a small amount of information
about the image (such as a few low order moments).  The great appeal of
normalization is that it isolates the problem of finding the image modulo
the effect of viewing transformations, from the higher order problem of 
deciding which features of the image are needed for a specific classification
decision.  Intuitively, normalization is simply a systematic method for
transforming from observer--based to image--based coordinates; in the former
the image depends on the view, whereas in the latter the image is viewing
transformation independent.  From a mathematical viewpoint, our method 
consists of placing a set of constraints on the transformed image equal 
in number to the number of viewing transformation parameters, permitting 
one to solve either algebraically or numerically for the parameters of
a normalizing transformation.   Since the constraints are necessarily 
viewing transformation noninvariants, their construction is in general 
simpler than the direct construction of viewing transformation invariants.  

Let us begin our discussion with a quick review of the viewing transformations
of a planar object, since these transformations will be used as illustrations 
of our general methods. (For further details, and a bibliography, see the 
excellent recent book of Reiss [15].)   Under rigid 3D motions the image 
$I(\vec x)$, with $\vec x=(x_1,~x_2)$ the two dimensional coordinate in the
image plane, is transformed to $I(\vec x^{\, \prime})$, with 
$\vec x^{\, \prime}$ related to
$\vec x$ by the {\it planar projective transformation} 
$$x_n^{\prime} = {\sum_{m=1}^2 ~G_{nm} ~x_m + t_n \over
1+\sum_{m=1}^2 ~p_m ~x_m}, ~~n=1,2 ~~.
 ~~\eqno(1)$$
When the depth of the object is much less than its distance from the lens, 
then the parameter $p_n$ in Eq.~(1) can be neglected, and Eq.~(1) reduces to
 the linear {\it affine} transformation
$$x_n^{\prime}= \sum_{m=1}^2 ~G_{nm} ~x_m ~+t_n~~. ~~\eqno(2)$$
[An affine transformation, with $G_{nm}$ replaced by $G_{nm}-t_n~p_m$, also
results when Eq.~(1) is expanded in a power series in $x_m$ and second and
higher order terms are neglected.]
Additionally, when the viewed object is constrained to lie in the plane normal
to the viewing or 3 axis, Eq.~(2) specializes further to the {\it similarity}
transformation group
of scalings, rotations, and translations, in which $G_{nm}$ is simply a
multiple (the scale factor) of a two dimensional rotation matrix.
The projective transformations, the affine transformations, and the similarity
transformations all form  groups, and this will be the characterizing feature
of the viewing transformations studied in our general analysis.

In applications, it will be convenient to use subgroup factorizations, which
are readily obtained from the group multiplication rule for the 
transformations
of Eqs.~(1) and (2).  For example, a general planar projective transformation
can be written as the result of composing what we will term a {\it restricted 
projective transformation}
$$x^{\prime \prime}_n= {x^{\prime}_n \over 1+\sum_{m=1}^2 ~p_m ~x^{\prime}_m }
\eqno(3)$$
with the general affine transformation of Eq.~(2).
Another subgroup factorization expresses the general affine transformation 
of Eq.~(2) as the result of the composition of a pure translation
$$x^{\prime \prime}_n=x^{\prime}_n + t_n  \eqno(4a)$$
with a homogeneous affine transformation 
$$x^{\prime}_n = \sum_{m=1}^2 ~G_{nm} ~x_m ~~. \eqno(4b)$$
Yet a third subgroup factorization expresses a general 
homogeneous affine transformation as the result
of composing  what we will term a {\it restricted affine transformation}, 
which
has vanishing upper right diagonal matrix element,
$$x^{\prime \prime}_n=\sum_{m=1}^2 ~g_{nm} ~x^{\prime}_m~, ~~~~g_{12}=0 ~~, 
\eqno(5a) $$
with a pure rotation
$$x^{\prime}_n=\sum_{m=1}^2 R_{nm} x_m~, ~~~~R_{11}=R_{22}= \cos \theta,
~~~~R_{12}=-R_{21} =-\sin \theta  ~~. \eqno(5b)$$
A variant of Eqs.~(5a--b) is obtained by requiring
that the matrix $g$ have unit determinant, so that it has the two--parameter
form $g_{11}=u,~g_{12}=0,~g_{21}=w,~g_{22}=u^{-1}$, and then including
a scale factor $\lambda$ in Eq.~(5b), which now reads
$$x^{\prime}_n=\lambda ~\sum_{m=1}^2 R_{nm} x_m. \eqno(5c)$$
  
\bigskip
\centerline{\bf 2.~~General Theory of Image Normalization }

We proceed now to formulate a general framework for image normalization, with  

the aim of understanding the common elements of the various normalization 
methods which appear in the literature and of generalizing them to new
applications.
As a preliminary to the mathematical discussion of Subsecs.~2A-E, we specify 
our
notation for viewing transformations.  Let 
${\cal G}=\{S\}$ be a group of symmetry or viewing 
transformations $S$, which act on the
image $I(\vec x)$ according to
$$I(\vec x) \to I_S(\vec x) = I(\vec S(\vec x)) ~~. \eqno(6a) $$
Our notational convention, that we shall adhere to throughout, is that 
$\vec x^{\prime}=\vec S(\vec x)$ is the concrete image coordinate 
mapping induced
by the abstract group element $S$.  [A specific example of such a 
transformation 
would be the planar projection transformation of Eq.~(1), in which $S$   
would be the abstract element of the planar projective group characterized 
by  the parameters $G_{mn},~t_n,~p_m$ specifying 
the concrete coordinate mapping.]  In this notation, the
result of successive transformations with $S_1$ followed by $S_2$ is given
by 
$$I(\vec x) \to I_{S_2 S_1}(\vec x)= I(\vec S_2(\vec S_1(\vec x))) ~~.
\eqno(6b) $$

The transformation groups of interest to us are in general ones with 
continuous parameters, in other words, Lie groups, and the reader interested 
in more background on Lie group theory may wish to consult the 
texts of Gilmore [8] 
and Sattinger and Weaver [13].  However, very little of the formal apparatus 
of Lie group theory is required in what follows; basically, all we use 
is the group closure property and the enumeration of the number 
of group parameters.  In particular, no knowledge of the 
representation theory of Lie groups is needed.  

\parindent=0pt
A.  {\it The normalization recipe}.  We begin by giving the general 
prescription for an image normalization transformation.  Let 
$\vec N_I(\vec x)$
be a transformation of $\vec x$  which depends on the image $I$, and
which is constructed so that under the image transformation of 
Eq.~(6a), it behaves as
$$\vec N_{I_S}(\vec x)=\vec S^{-1}(\vec N_I(\vec x))  ~~, \eqno(7a)$$
with $\vec S^{-1}$ the inverse transformation to $\vec S$ of Eq.~(6a),
$$\vec S(\vec S^{-1}(\vec x))=\vec x  ~~. \eqno(7b)$$
Also, let $\vec M_I(\vec x)$ be an optional second transformation 
of $\vec x$ which depends
on the image $I$ only through invariants under the group of transformations
$\cal G$, that is, 
$$\vec M_{I_S}(\vec x) = \vec M_I(\vec x), ~~~~~ {\rm all} \> S \in \cal G ~~.
\eqno(7c)$$
Then 
$$\tilde I(\vec x) =I(\vec N_I(\vec M_I(\vec x))) \eqno (8) $$
is a normalized image which is invariant under all transformations of the
group $\cal G$.  This is an immediate consequence of Eq.~(6a) and 
Eqs.~(7a--c), from which we
have
$$\eqalign{
\tilde I_S(\vec x)=&I_S(\vec N_{I_S}(\vec M_{I_S}(\vec x))) \cr
=&I(\vec S(\vec S^{-1}(\vec N_I(\vec M_I (\vec x))))) \cr
=&I(\vec N_I(\vec M_I(\vec x))) = \tilde I(\vec x)~~. \cr
}\eqno(9)$$
 
\parindent=0pt
B.  {\it Uniqueness}.  Before specifying how to actually construct a map
$\vec N_I$ obeying Eq.~(7a), let us address the issue of uniqueness.  That
is, given {\it two} maps $\vec N_{1I}(\vec x)$ and $\vec N_{2I}(\vec x)$,
both of which obey Eq.~(7a), how are they related?  By hypothesis, we have
$$\eqalign{
\vec N_{1I_S}(\vec x)=&\vec S^{-1}(\vec N_{1I}(\vec x)) ~~, \cr
\vec N_{2I_S}(\vec x)=&\vec S^{-1}(\vec N_{2I}(\vec x)) ~~. \cr
}\eqno(10)$$
Since for any $\vec f(\vec x)$ and $\vec g(\vec x)$ we have
$$\vec f(\vec g(\vec x))^{\, -1}=\vec g^{~-1}(\vec f^{-1}(\vec x)) ~~, 
\eqno (11a)$$
we can rewrite the first line of Eq.~(10) as
$$\vec N_{1I_S}^{-1}(\vec x)=\vec N_{1I}^{-1}(\vec S(\vec x)) ~~. \eqno 
(11b)$$
Let us now define a new map $\vec M_I(\vec x)$ by
$$\vec M_I(\vec x) \equiv \vec N_{1I}^{-1}(\vec N_{2I}(\vec x)) ~~, \eqno 
(12)$$
which reduces to the identity map when $\vec N_{1I}=\vec N_{2I}$; 
then by Eq.~(11b) and the first line of Eq.~(10), we have
$$\eqalign{
\vec M_{I_S}(\vec x)=& \vec N_{1I_S}^{-1}(\vec N_{2I_S}(\vec x)) \cr
=&\vec N_{1I}^{-1}(\vec S(\vec S^{-1}(\vec N_{2I}(\vec x))))  \cr
=&\vec N_{1I}^{-1}(\vec N_{2I}(\vec x))= M_I(\vec x)  ~~. \cr
}\eqno (13a)$$
In other words, $M_I(\vec x)$ depends on the image $I$ only through invariants
under transformations of the group $\cal G$, and from Eq.~(12), the 
normalizing
map $\vec N_{2I}$ is related to the normalizing map $\vec N_{1I}$ by
$$\vec N_{2I}(\vec x) = \vec N_{1I}(\vec M_I(\vec x))~~. \eqno(13b)$$
This is why in writing the general normalized image corresponding to a
particular normalizing map in Eq.~(8), we have included in the $\vec x$ 
dependence
the possible appearance of a map $\vec M_I$ which depends on the
image only through invariants under transformation by elements of $\cal G$.  

\parindent=0pt
C.  {\it Construction of $\vec N_I$ by imposing constraints, and demonstration
that normalization yields a complete set of invariants}.  We next
show that one can construct an image normalization transformation obeying
Eq.~(7a) by imposing a suitable set of constraints.  We shall assume now
that $\cal G$ is a $K$--parameter Lie group which is continuously connected to 

the identity.  Let $C_k[I]=C_k[I(\vec x)]~,~~~k=1,...,K$ 
(where $\vec x$ is a dummy 
variable)  be a set of functionals of the
image $I(\vec x)$ with the property that the $K$ constraints 
$$C_k[I_{S^{\prime}}]=
C_k[I(\vec S^{\prime}(\vec x))]=0~,~~~k=1,...,K  \eqno(14a)$$
are satisfied for a unique element $S^{\prime}=N_I$ of $\cal G$, so that
$$C_k[I(\vec N_I(\vec x))]=0~,~~~k=1,...,K~.  \eqno(14b)$$
Then, as we shall now show,  
$\vec N_I(\vec x)$ is the desired normalizing transformation.

\parindent=25pt
We remark that the condition that Eqs.~(14a, b) should  
have a unique solution can be relaxed in applications to the condition  
that there be only one solution in the range of 
relevant viewing transformation 
parameters.   Clearly, either form of the uniqueness condition 
requires that the constraint functionals {\it not} be 
invariants under $\cal G$, and thus their structure will in general be 
simpler than that of directly constructed viewing transformation invariants.   
  
In many cases, as we will see in Sec. 3 below, the constraints 
can be constructed from viewing transformation {\it covariants}, which 
have simple algebraic properties under the transformations of $\cal G$, 
permitting closed form algebraic solution for the parameters of the 
normalizing transformation.  In more complicated cases, as discussed 
in Sec. 4, the constraints must be solved numerically for the normalizing 
transformation. 

\parindent=25pt
To see that the construction of Eqs. (14a,b) gives a transformation 
$\vec N_I(\vec x)$ that 
obeys Eq.~(7a), let us consider the effect of replacing
$I$ by $I_S$ in Eqs.~(14a, b).  By hypothesis, the constraints
$$C_k[I_S(\vec S^{\prime}(\vec x))]=0~,~~~k=1,...,K \eqno(15a)  $$
are uniquely satisfied by a group element $S^{\prime}=N_{I_S}$ of $\cal G$, so 
that
$$C_k[I_S(\vec N_{I_S}(\vec x))]=0~,~~~k=1,...,K~, \eqno(15b)~  $$
with $\vec N_{I_S}(\vec x)$ 
the proposed normalizing transformation corresponding to $I_S$.  But using  
\break 
Eq.~(6a), we can also write Eq.~(15b) as
$$C_k[I(\vec S(\vec N_{I_S}(\vec x)))]=0~,~~~k=1,...,K~, \eqno(15c)  $$
which has the same structure as Eq.~(14b). Therefore, 
by uniqueness of the solution
$N_I$ of Eq.~(14b) we must have
$$\vec S(\vec N_{I_S}(\vec x))=\vec N_I (\vec x) ~,  \eqno(16a)$$
which by Eq.~(7b) is equivalent to
$$\vec N_{I_S}(\vec x)=\vec S^{-1}(\vec N_I(\vec x)) ~, \eqno (16b) $$
showing that the $N_I$ produced by solving the constraints does indeed
obey Eq.~(7a).  Hence the imposition of constraints gives a constructive
procedure for generating image normalization transformations.  
\parindent=25pt

We note that
this construction makes the normalizing transformation $\vec N_I$ an element 
of
the group $\cal G$, and the quotient $\vec M_I(\vec x) \equiv 
\vec N_{1I}^{-1}(\vec N_{2I}(\vec x))$ of two normalizing maps constructed
by imposing different sets of constraints will likewise be an element of 
$\cal G$. When both $\vec N_I$ and $\vec M_I$ in Eq.~(8)
belong to $\cal G$, we can invert Eq.~(8) to express the original image $I$
in terms of the invariant, normalized image $\tilde I$ according to
$$I(\vec x)=\tilde I(\vec M_I^{-1}(\vec N_I^{-1}(\vec x)))~. \eqno(16c) $$
This equation shows that normalization leads to a complete set of invariants, 
in the sense that the information in the normalized image, plus the
$K$ parameters determining the viewing transformation 
$\vec M_I^{-1}(\vec N_I^{-1}(\vec x))$, suffice to completely reconstruct the
original image.  By way of contrast, the representation--theoretic methods
discussed in Sec. 6.5 of Lenz [12], and the integral transform methods
of Ferraro [7], although attacking the same problem
as is discussed here, yield only a small fraction of the complete set of
invariants.  Moreover, normalization has the further advantage of requiring 
only a minimal knowledge of 
the kinematic structure of the group; the full
irreducible representation structure is not needed, and the methods described
here are applicable to noncompact as well as to compact groups.  
We note finally that
the discussion of this section is slightly less
general than that of Secs.~3A and 3B, where we did not require either $\vec 
N_I$
or $\vec M_I$ to belong to $\cal G$; the most general normalizing map
$\vec N_I$ is obtained from one generated by constraints by using as its
argument a map $\vec M_I$ which does not belong to $\cal G$ but that is   
invariant under transformations of the image $I$ by $\cal G$.

\parindent=0pt
D.  {\it Extension to reflections and contrast invariance.}  We consider next
two simple extensions of the constraint method for constructing the 
normalizing
transformation.  The first involves relaxing the requirement that $\cal G$ be 
simply connected to the identity, as is needed if $\cal G$ contains improper
transformations such as reflections.  Reflections are said to be 
independent if they 
do not differ solely by an element of the connected component of the group; 
for each independent 
discrete reflection $R$ in $\cal G$, the set of constraints of Eq.~(14a) 
must be augmented by
an additional constraint $D[I(\vec S^{\prime}(\vec x))] > 0$, where 
$D[I(\vec x)]$
is a functional of the image which changes sign under the reflection operation 

$R$, 
$$D[I(\vec R(\vec x))]=-D[I(\vec x)] ~. \eqno(17)$$
The second extension involves incorporating invariance under changes of
image contrast, that is, under image transformations of the form
$$I(\vec x) \to cI(\vec x), ~~~~~ c >0~.  \eqno(18a) $$
To the extent that illumination is sufficiently slowly varying that it
can be treated as constant over a viewed object, changes in illumination
level as the object is moved to different views take the form of changes
in the constant $c$ in Eq.~(18a), which is why incorporating contrast
invariance can be important.  
If we require that the constraint functionals $C_k$ [and $D$ if needed]
should be invariant under the change of contrast of Eq.~(18a), then
the image normalization transformation $\vec N_I(\vec x)$ and the
auxiliary transformation $\vec M_I(\vec x)$ can be taken to be contrast
invariant.  A contrast invariant normalized image $\tilde I_c(\vec x)$
is then obtained by the obvious recipe
$$\tilde I_c(\vec x)= {\tilde I(\vec x) \over \int ~d^2x \tilde I(\vec x)} ~.
\eqno (18b) $$

\parindent=0pt
E.  {\it Use of subgroup decompositions.}  Suppose that for a general
element $S$ of the group $\cal G$, there is a subgroup decomposition of the 
form
$$S=S_2S_1~,   \eqno(19a)$$
with $S_2$ belonging to a subgroup ${\cal G}_2$ of $\cal G$, $S_1$ belonging 
to
a subgroup ${\cal G}_1$ of $\cal G$, and with the respective parameter counts 
$K, K_1$, and $K_2$ of $\cal G$,~${\cal G}_1$, and ${\cal G}_2$ obeying
$$K=K_1+K_2 ~. \eqno(19b)$$
(Such subgroup compositions for a general Lie group are obtained by 
constructing a composition series for the group, but we will not need 
this formal apparatus in the relatively simple applications that follow.)
Let us suppose further that we can solve 
the problem of image normalization with respect to the group ${\cal G}_1$, 
and that we wish to extend this solution to the full invariance group $\cal 
G$.
The subgroup decomposition allows this to be done by imposing $K_2$ additional
constraints to deal with the ${\cal G}_2$ subgroup, as follows.  
Let $C_{2k}[I(\vec x)]$,
with $k=1,...,K_2$, be a set of functionals of the image chosen so that
the constraints
$$C_{2k}[I(\vec N_{2I}(\vec S_1 (\vec x)))]=0~, ~~~k=1,...,K_2  \eqno(20a)$$
are independent of $S_1 \in {\cal G}_1$. 
In particular, taking $S_1$ as the identity transformation, Eq.~(20a) 
simplifies
to
$$C_{2k}[I(\vec N_{2I}(\vec x))]=0~, ~~~k=1,...,K_2~,  \eqno(20b)$$
which if we impose the requirement of a unique solution over transformations
$N_2 \in {\cal G}_2$ determines a ``partial normalization'' transformation 
$\vec N_{2I}$.
Note that a sufficient condition for the constraints of Eq.~(20a) 
to be independent of $S_1$ is for the
functionals $C_{2k}$ to be $S_1$--independent, 
but this is not a necessary condition;
we will see examples in which, as $S_1$ traverses ${\cal G}_1$, 
the functionals are merely covariant in some
simple way that guarantees invariance of the constraints obtained by 
equating all the functionals to zero.  To see how $\vec N_{2I}$ transforms
under the action of the group $\cal G$, we replace $I$ by $I_S$ in Eq.~(20b),
giving  
$$C_{2k}[I_S(\vec N_{2I_S}(\vec x))]=0~, ~~~k=1,...,K_2~~~;  \eqno(21a)$$
again making use of Eq.~(6a) this becomes
$$C_{2k}[I(\vec S(\vec N_{2I_S}(\vec x)))]=0~, ~~~k=1,...,K_2~~~.  
\eqno(21b)$$
Since the argument $\vec S(\vec N_{2I_S}(\vec x))$ appearing in Eq.~(21b) 
is no longer a member of
the ${\cal G}_2$ subgroup, we cannot conclude that it is equal to the argument
$\vec N_{2I}(\vec x)$ appearing in Eq.~(20b), but the arguments can differ
at most by a transformation of $\vec x$ by some member $\vec S_1^{\prime}$ of 
the
subgroup ${\cal G}_1$ which leaves the constraints invariant, giving
$$\vec N_{2I_S}(\vec x)=\vec S^{-1}(\vec N_{2I}(\vec S_1^{\prime}
(\vec x)))
~~~ \eqno(22a)$$
as the subgroup analog of Eq.~(7a).  Corresponding to this, the partially
normalized image defined by
$$\tilde I(\vec x) =I(\vec N_{2I}(\vec x))   \eqno(22b)$$
transforms under the group $\cal G$ as
$$\eqalign{
\tilde I(\vec x) \to \tilde I_S(\vec x)=
&I_S(\vec N_{2I_S}(\vec x))  \cr
=&I(\vec S(\vec S^{-1}(\vec N_{2I}(\vec S_1^{\prime}(\vec x))))) \cr
=&I(\vec N_{2I}(\vec S_1^{\prime}(\vec x)))=\tilde I
(\vec S_1^{\prime}(\vec x)) ~, \cr
}\eqno(22c) $$
and thus changes only by a transformation lying in the ${\cal G}_1$ subgroup.  

Further image normalization of $\tilde I$ 
using the constraints appropriate to ${\cal G}_1$ then
gives a final normalized image
$$ \hat I(\vec x) = I(\vec N_{2I}(\vec N_{1 \tilde I}
(\vec M_{I}(\vec x))))~, \eqno(23)$$
which is invariant with respect to the full group of transformations $\cal G$,
where as before $\vec M_{I}$ is any transformation which is constructed solely
using $\cal G$ invariants of the image.
\bigskip
\centerline{\bf 3.~~Viewing Transformations of a Planar Object} 
\centerline{\bf With Algebraically  Solvable Constraints}
\parindent=25pt

We proceed now to apply the general image normalization 
methods of Sec.~2 to the
viewing transformations of a planar object.   In this section we 
focus on cases, corresponding to linear viewing transformations,  
in which suitable constraints can be formed using simple 
viewing transformation covariants, leading to algebraically solvable 
constraints.  In the next section we will discuss more complicated cases, 
several of which use the transformations of this section as building blocks, 
in which some of the constraints must be solved by iterative methods.    

\parindent=0pt
A.  {\it Translations}.  The translation subgroup of Eq.~(1) is given by
$$\vec S(\vec x)=\vec x + \vec t ~, \eqno(24a) $$
corresponding to which $I_S=I(\vec x +\vec t\,)$ describes an 
image translated by the vector
$-\vec t$.  We take as the constraint functionals 
$$C_k[I_S]=\int ~d^2x ~x_k ~I(\vec x +\vec t\,)
,~~~~ k=1,2~. \eqno(24b) $$
The constraints $C_k=0,~~~k=1,2$ can be solved explicitly for $\vec t$ 
by making the change of integration variable $\vec y=\vec x + \vec t$, 
giving the unique solution $\vec t = \vec t_I$, with $\vec t_I$ the image
``center of mass''
$$\vec t_I= {\int ~d^2x ~\vec x ~I(\vec x) \over \int ~d^2x ~I(\vec x)} ~,
\eqno(25a)$$
and the corresponding normalizing transformation is
$$\vec N_I(\vec x)=\vec x+ \vec t_I ~. \eqno(25b)$$
Under the action of the translation $S$, Eq.~(25a) becomes
$$\vec t_{I_S}= {\int ~d^2x ~\vec x ~I(\vec x +\vec t\,) \over 
\int ~d^2x ~I(\vec x +\vec t\,)} ~,
\eqno(26a)$$
which by a change of integration variable yields
$$\vec t_{I_S}=\vec t_I -\vec t ~. \eqno(26b)$$
Thus the normalizing transformation of Eq.~(25b) behaves as
$$\vec N_{I_S}(\vec x)=\vec x+ \vec t_I -\vec t = \vec S^{-1}(\vec N_I(\vec 
x))
~,  \eqno(26c)$$
in agreement with the general result of Eq.~(7a).  In accordance with
Eq.~(8), the translation invariant image is
$$\tilde I(\vec x)=I(\vec N_I(\vec M_I(\vec x)))=
I(\vec M_I(\vec x) + \vec t_I)~,
\eqno(27a)$$
with $\vec M_I(\vec x)$ any transformation of $\vec x$ which depends only on
translation invariant image features.  Usually, one makes the choice
$\vec M_I(\vec x) =\vec x - \vec t_0$, with $\vec t_0$ a constant vector
which is independent of the image $I$.  This constant vector can of course 
be taken to be zero, corresponding to the choice
$$\vec M_I(\vec x) =\vec x ~, \eqno(27b)$$
or it can be adjusted to center the
translation invariant form of {\it one particular} image $I_0$ at any desired
point.  
\parindent=25pt

Once we have the translation normalized image $\tilde I(\vec x)$, all features
extracted from it, such as all Fourier transform or wavelet transform
amplitudes, are translation invariant.  We illustrate this explicitly in the
case of the Fourier transform, by showing how a translation invariant
Fourier transform $\tilde I(\vec k)$ 
is related to the Fourier transform $I(\vec k)$ of the original 
image $I(\vec x)$,
$$\eqalign{
\tilde I(\vec k)=&\int ~d^2x e^{-i \vec k \cdot \vec x} \tilde I(\vec x) \cr
=&\int ~d^2x e^{-i \vec k \cdot \vec x} I(\vec x+ \vec t_I) \cr
=&e^{i \vec k \cdot \vec t_I} ~I(\vec k), \cr
} \eqno(28a)$$
where we have taken $\vec M_I$ to be the identity map as in Eq.~(27b), and
where 
$$I(\vec k)=\int ~d^2x e^{-i \vec k \cdot \vec x} I(\vec x)~. \eqno(28b)$$
Under translation, the Fourier transform of the original image $I(\vec k)$
behaves as
$$I(\vec k) \to I_S(\vec k) =\int ~d^2x e^{-i \vec k \cdot \vec x} 
I(\vec x+ \vec t)
=e^{i \vec k \cdot \vec t} I(\vec k) ~, \eqno(28c)$$
and is not invariant, but by Eq.~(28b), the factor $e^{i \vec k \cdot \vec 
t_I}$
behaves as
$$e^{i \vec k \cdot \vec t_I} \to e^{i \vec k \cdot \vec t_{I_S}}
=e^{-i \vec k \cdot \vec t} e^{i \vec k \cdot \vec t_I}~, \eqno(28d)$$
and has a compensating noninvariance, making the product $\tilde I(\vec k)$
appearing in Eq.~(28a) invariant under image translations.
\parindent=0pt

B.  {\it Separation of affine normalization into translational and homogeneous
affine normalization.}  Since rotations and scalings are special cases of
homogeneous affine transformations, before discussing them we use the subgroup
decomposition method to give the general procedure for separating the
affine normalization problem into a translational normalization followed
by a homogeneous affine normalization. We follow the general procedure of
Eqs.~(19a)--(23), taking the subgroup ${\cal G}_2$ to be the translations, and 
the
subgroup ${\cal G}_1$ to be the homogeneous affine 
transformations, as in Eqs.~(4a, b), so that
$$\eqalign{
&\vec S_2(\vec x) =\vec x +\vec t~,~~~~\vec S_1(\vec x)=G \cdot \vec x~, \cr
&\vec S(\vec x)=\vec S_2(\vec S_1(\vec x))=G \cdot \vec x + \vec t~,  \cr
}\eqno(29a)$$
where the notation $G \cdot\vec x$ denotes the vector with components
$\sum_{m=1}^2 G_{nm} ~x_m$.  Applying the same translational constraint 
functionals
$C_k$ of Eq.~(24b) to the general affine transformation $\vec S$ of Eq.~(29a),
we have 
$$C_k[I_S]=\int ~d^2x ~x_k ~I(G \cdot\vec x + \vec t\,)
~, \eqno(29b)$$
which on making the change of integration variable 
$$\vec x \to \vec S_1^{-1}(\vec x) = G^{-1} \cdot \vec x~, \eqno(29c)$$
gives 
$$C_k[I_S]=\sum_{\ell} (G^{-1})_{k \ell} \int ~d^2x ~x_{\ell} ~I(\vec x
+\vec t\,) 
= \sum_{\ell} (G^{-1})_{k \ell} C_{\ell}[I_{S_2}]~.  
\eqno(29d)$$
Thus, although the translational constraint {\it functionals} $C_k$ are 
not independent of $S_1$ (or $G$), they simply mix linearly 
when $S_1$ is changed,
and consequently the translational {\it constraints}
$C_k=0~,~~k=1,2$ are $S_1$--independent.  This permits us to normalize out the
translational part of a general affine transformation independently of the
homogeneous affine transformations, leading to a partially normalized
image
$$\tilde I(\vec x) =I(\vec x +\vec t_I) \eqno(30a)$$
which is translation invariant.  Under the full affine group, $\vec t_I$
transforms as
$$\vec t_I \to \vec t_{I_S} ={\int ~d^2x ~\vec x ~I(G \cdot \vec x +\vec t\,) 
\over \int ~d^2x ~I(G \cdot \vec x +\vec t\,)}~, \eqno(30b)$$
which by the same changes of integration variable used before reduces to
$$\vec t_{I_S}=G^{-1} \cdot (\vec t_I -\vec t~)~.\eqno(30c)$$
Hence the partially normalized image of Eq.~(30a)
transforms under the full affine group as
$$\eqalign{
\tilde I(\vec x) \to \tilde I_S(\vec x)=&
I_S(\vec x + \vec t_{I_S})  \cr
=&I(G \cdot [x+ G^{-1} \cdot (\vec t_I -\vec t\,)] +\vec t~)
=I(G \cdot \vec x +\vec t_I\,)
\cr =&\tilde I(G \cdot \vec x) ~, \cr
}\eqno(31a) $$
in agreement with the general result of Eq.~(22c).  In other words, the
partially normalized image $\tilde I$ is translation invariant, and is
acted on only by the homogeneous part of the affine transformation.  
In discussing similarity and affine transformations in Subsecs. C--F which
follow, we will assume that we are always dealing with a partially normalized
image which is translation invariant, but for simplicity of notation we
will drop the tilde and simply call this image $I$.  However, keeping the 
tilde
for the moment, we note that the moments $\mu_{pq}$ 
of this partially normalized
image, which are called {\it central moments}, are defined by 
$$\mu_{pq}=\int_{-\infty}^{\infty}~dx_1 \int_{-\infty}^{\infty}~dx_2
~~x_1^p ~x_2^q ~\tilde I(\vec x) ~~.\eqno(31b)$$
\parindent=0pt

C.  {\it Rotations.}  We begin the discussion of homogeneous 
affine transformations
by considering pure rotations, with the group action 
$$\vec S(\vec x)=\vec x_{\theta} \equiv (x_1 \cos \theta - x_2 \sin \theta,
x_1 \sin \theta + x_2 \cos \theta) ~, \eqno(32)$$
corresponding to which 
$I_S(\vec x) =I(\vec x_{\theta})$ describes an image rotated by the
angle $-\theta$.  We shall assume, for the moment, that the image to be
normalized has no rotational symmetry, in which case we can take as the
constraint functional
$$C[I_S]={\rm Phase}\left[\int ~d^2x ~e^{i \Phi(\vec x~)} ~f(|\vec x|)
 ~I(\vec x_{\theta})\right] -1 ~. \eqno(33a)$$
Here we have used the notation
$${\rm Phase}[z]=z/|z|  \eqno(33b) $$
for the complex number $z$; the function $f$ is arbitrary [6] , 
and the functions $\Phi(\vec x~)$
and $|\vec x|$ are defined by
$$\Phi(\vec x~) \equiv \arctan(x_2/x_1)~, ~~~|\vec x| \equiv 
\sqrt{x_1^2+x_2^2} ~.
\eqno(33c)  $$
The constraint $C[I_S]=0$ now uniquely determines an angle $\theta=
\theta_I$, which can be calculated explicitly by making a change of variable
$\vec x \to \vec x_{-\theta}$ in Eq.~(33a) and using the trigonometric formula
$$\Phi(\vec x_{- \theta})=\arctan({-x_1 \sin \theta + x_2 \cos \theta
\over x_1 \cos \theta +x_2 \sin \theta})=\Phi(\vec x) - \theta ~, \eqno(34a)$$ 

thus giving
$$e^{i \theta_I} = {\rm Phase}\left[\int ~d^2x ~e^{i \Phi(\vec x)} 
~f(|\vec x|) ~I(\vec x)\right] ~,  \eqno(34b)$$
with the corresponding normalization transformation
$$\vec N_I(\vec x)=\vec x_{\theta_I}=(x_1~\cos ~\theta_I - x_2 
~\sin \theta_I, x_1 ~\sin \theta_I + x_2 \cos \theta_I) ~. 
\eqno(34c)  $$
Under the action of the rotation $S$, Eq.~(34b) is transformed to
$$e^{i \theta_{I_S}} = {\rm Phase}\left[\int ~d^2x ~e^{i \Phi(\vec x)} 
~f(|\vec x|) ~I(\vec x_{\theta})\right] ~,  \eqno(35a)$$
which making the change of variable $\vec x \to \vec x_{-\theta}$ and using
Eq.~(34a) gives
$$\theta_{I_S}=\theta_I-\theta ~. \eqno(35b)$$
Thus the normalization transformation $\vec N_I$ becomes, under the action
of $S$,
$$\vec N_{I_S}(\vec x)=\vec x_{\theta_{I_S}}=\vec x_{\theta_I-\theta}
=\vec S^{-1}(\vec N_I(\vec x)) ~,
\eqno(35c)$$
in agreement with Eq.~(7a).  Following the prescription of Eq.~(8), the
rotationally normalized image is
$$\tilde I(\vec x) = I(\vec N_I(\vec M_I(\vec x)))
=I(\vec M_I(\vec x)_{\theta_I})  ~, \eqno(36)$$
with $\vec M_I(\vec x)$ any transformation of $\vec x$ which depends only on
rotationally invariant image features.  Usually, one makes the choice
$\vec M_I(\vec x) = \vec x_{\theta_0}$, with $\theta_0$ a constant angle
which is independent of the image $I$.  This angle can of course be taken 
to be zero, corresponding to the $\vec M_I$ of Eq.~(27b), or it
can be used to give the rotationally invariant form of one particular image
a specified orientation.  Note that the angle $\theta_I$ used to construct
the normalizing transformation contains useful information about the 
orientation of the image in the observer--centered coordinate system, which
can be used to disambiguate images which have the same invariant form, but
a different classification depending on their absolute orientation.  For
example, in some typefaces a $6$ and a $9$ have the same rotationally
normalized form, but their $\theta_I$ values will differ by $\pi$, and so
the value of $\theta_I$ modulo $2 \pi$ can be used to resolve the six--nine
ambiguity.  
\parindent=25pt

Up to this point we have assumed that the image to be normalized has no
special rotational symmetries.  Suppose now that $I$ has an $N-$fold
rotational symmetry, so that $I(\vec x)=I(\vec x_{2 \pi/N})$, and let us
consider the integral 
$$\eqalign{
&\int ~d^2x ~e^{i M \Phi(\vec x)} ~f(|\vec x|) ~I(\vec x)=\int ~d^2x
~e^{i M \Phi(\vec x)} ~f(|\vec x|) I(\vec x_{2 \pi/N}) \cr
&=\int ~d^2x ~e^{i M \Phi(\vec x_{-2 \pi/N})} ~f(|\vec x|) ~I(\vec x)  
=e^{-i M 2 \pi/N} ~\int ~d^2x ~e^{i M \Phi(\vec x)} ~f(|\vec x|) 
~I(\vec x) ~, \cr
} \eqno(37a) $$
which implies that the integral
$$\int ~d^2x ~e^{i M \Phi(\vec x)} f(|\vec x|) ~I(\vec x)   \eqno(37b)$$
vanishes unless $M/N$ is an integer (Hu [11]; 
Abu--Mostafa and Psaltis [2]).  Consequently, when there is a rotational
symmetry, the constraint of Eq.~(33a) can no longer be used; instead, we must
find the smallest value  $M=N$ for which the integral of Eq.~(37b) is 
nonvanishing, and then normalize using the constraint functional $C_N[I_S]$
defined by
$$C_N[I_S]={\rm Phase}\left[\int ~d^2x ~e^{i N \Phi(\vec x)} ~f(|\vec x|)
 ~I(\vec x_{\theta})\right] -1 ~. \eqno(38a)$$
Solving the constraint $C_N[I_S]=0$ now determines an angle $\theta =
\theta_I$, unique up to an integer multiple of $2 \pi/N$, which can be
explicitly calculated from 
$$e^{i N \theta_I} = {\rm Phase}\left[\int ~d^2x ~e^{i  N \Phi(\vec x)} 
~f(|\vec x|) ~I(\vec x)\right] ~,  \eqno(38b)$$
with the corresponding normalizing transformation and normalized image
still given by Eqs.~(34c) and (36).  In practice, one does not know {\it a 
priori} the rotational symmetry of the image being normalized; one then 
deals with the possibility
of rotational symmetry by taking the constraint functional to have the form of
Eq.~(38a) and including a loop over $N=1,2,...$ which terminates at the
smallest value of $N$ for which the integral used to construct the constraint 
is nonvanishing.
\parindent=0pt

D.  {\it Scaling.}  We turn next to scaling, with the group action
$$\vec S(\vec x)=\lambda \vec x~, ~~~~\lambda >0  ~, \eqno(39)$$
corresponding to which $I_S(\vec x)=I(\lambda \vec x)$ describes 
an image scaled in
size by a factor $\lambda^{-1}$.  We take as the constraint functional
$C_{\mu \nu}[I_S], ~~~~\mu \neq \nu$, given by
$$C_{\mu \nu}[I_S]={\int ~d^2x |\vec x|^{\mu} g(\Phi(\vec x)) 
I(\lambda \vec x) \over \int ~d^2x |\vec x|^{\nu} g(\Phi(\vec x)) 
I(\lambda \vec x)}-1 ~, \eqno(40a)$$
with $|\vec x|$ and $\Phi(\vec x)$ as in Eq.~(33c), and with $g$ an arbitrary
function.  The constraint $C_{\mu \nu}
=0$ determines a unique solution $\lambda=\lambda_I$, which by making the
change of integration variable $\vec x \to \vec x/\lambda$ is readily found
to be
$$\lambda_I=\left[{\int ~d^2x |\vec x|^{\mu} g(\Phi(\vec x)) 
I(\vec x) \over \int ~d^2x |\vec x|^{\nu} g(\Phi(\vec x)) 
I(\vec x)}\right]^{{1 \over \mu-\nu}}   ~. \eqno(40b)$$
The scale normalization transformation $\vec N_I(\vec x)$ is then
constructed as
$$\vec N_I(\vec x)=\lambda_I \vec x ~, \eqno(40c)$$
and under the action of the scaling $S$, is transformed to
$$\vec N_{I_S}(\vec x)=\lambda_{I_S} \vec x ~. \eqno (40d)$$ 
Writing the analog of Eq.~(40b) for $\lambda_{I_S}$, substituting Eq.~(39) 
and scaling 
$\lambda$ out of the integration variable as above, a simple calculation gives
$$\lambda_{I_S}=\lambda^{-1} \lambda_I ~, \eqno(41a)$$
and so the normalization transformation $N_I$ becomes, under the action of 
$S$,
$$N_{I_S}(\vec x) =\lambda^{-1} \vec N_I(\vec x)= 
\vec S^{-1}(\vec N_I(\vec x))  ~, \eqno(41b)$$
again in agreement with Eq.~(7a).  Following the recipe of Eq.~(8), the
image normalized with respect to scaling is
$$\tilde I(\vec x)=I(\vec N_I(\vec  M_I(\vec x))) ~, \eqno(42a)$$
with $\vec M_I(\vec x)$ any transformation of $\vec x$ which depends only on
scaling invariant image features.  The customary choice of $\vec M_I$ is
$\vec M_I(\vec x)= \lambda_0 \vec x$, with $\lambda_0$ an image--independent
scale factor, which can be used to make the normalized form of one particular
image have a specified size.  With this choice of $\vec M_I$, 
the normalizing integral
$\int ~d^2x \tilde I(\vec x)$ appearing in the contrast normalized image
of Eq.~(18b) can be explicitly calculated in terms of 
the central moment $\mu_{00}$, giving
$$\int ~d^2x \tilde I(\vec x)={ \mu_{00} \over (\lambda_I ~\lambda_0)^2}  ~.
\eqno(42b)$$
\parindent=25pt
 
A useful specialization of the general method for scaling normalization is
to take as the constraint
$$\eqalign{
0=&\left.{\partial \over \partial \mu} C_{\mu \nu}[I_S]\right\vert_{\mu=\nu=0} 
\cr
=&{\int ~d^2x ~\log |\vec x| ~g(\Phi(\vec x)) ~I(\lambda \vec x)  \over
\int ~d^2x ~g(\Phi(\vec x)) ~I(\lambda \vec x) } ~.\cr }  \eqno(43a) $$
By a change of integration variable this can be solved to give a
unique value $\lambda=\lambda_I$ given by
$$\log \lambda_I= {\int ~d^2x ~\log |\vec x| ~g(\Phi(\vec x)) ~I(\vec x)  
\over
\int ~d^2x ~g(\Phi(\vec x)) ~I(\vec x) } ~, \eqno(43b)$$
and which transforms under $S$ as
$$\log \lambda_{I_S}=\log(\lambda^{-1} \lambda_I) ~, \eqno(43c)$$
in agreement with Eq.~(41a).  A potential advantage of the logarithmic
weighting factor in Eq.~(43b), as compared with the power law weighting
factors in Eq.~(40a),  is that the logarithm does not suppress the 
contribution
arising from either the center or the periphery of the image.
\parindent=0pt

E.  {\it General similarity transformations.}  The general similarity 
transformation consists of a translation, a rotation, and a scaling, and so
can be normalized using the methods of Subsecs. 2A--D.  The rotational
and scaling normalizations can evidently be combined into a single step,
in which the arbitrary functions $f(|\vec x|)$ and $g(\Phi(\vec x))$ no longer
appear, having been replaced by the specific weightings appropriate to 
scaling and rotational normalizations.  In the image normalized
with respect to general similarity transformations, the undetermined 
map $\vec M_I(\vec x)$ can now depend only on general similarity invariants of 
the
image, and can of course be taken to be an image--independent map, including
the identity transformation.  
\parindent=0pt

F.  {\it Homogeneous affine transformations.}  We turn next to the general 
case of the homogeneous affine transformation, with the group action
$$\vec S(\vec x)=G \cdot \vec x \equiv (G_{11} x_1 + G_{12} x_2,
~G_{21} x_1 + G_{22} x_2) .  \eqno(44a) $$  
We again follow the subgroup
decomposition method of Eqs.~(19a)--(23), now taking (Dirilten and Newman [6]) 

the subgroup ${\cal G}_2$ to
be the group of affine transformations with vanishing upper diagonal matrix
element, and the subgroup ${\cal G}_1$ to be a rotation $R$, 
as in Eqs.~(5a,b), so
that
$$\eqalign{
&\vec S_2(\vec x)=g \cdot \vec x~, ~~~~\vec S_1(\vec x)=\vec x_{\theta}, \cr
&\vec S(\vec x)=\vec S_2(\vec S_1(\vec x))= g \cdot \vec x_{\theta}
=G \cdot \vec x~.  \cr
} \eqno(44b)$$
To normalize the image with respect to the three--parameter group ${\cal 
G}_2$, 
we will need three constraints, which following [15] we take as
$$C_k[I_S]=0~, ~~~~k=1,2,3~, \eqno(45a)$$
with the constraint functionals $C_{1,2,3}[I_S]$ given by 
$$\eqalign{
&C_1[I_S]={\int ~d^2x ~x_1^2 ~I(G \cdot \vec x)  \over \int ~d^2x 
~I(G \cdot \vec x) } -1   ~, \cr
&C_2[I_S]={\int ~d^2x ~x_2^2 ~I(G \cdot \vec x)  \over \int ~d^2x 
~I(G \cdot \vec x) } -1   ~, \cr
&C_3[I_S]=\int ~d^2x ~x_1 ~x_2 ~I(G \cdot \vec x)   ~. \cr
}\eqno(45b)$$
Although the constraint functionals of Eq.~(45b) 
are not independent of the element $S_1$ of the subgroup ${\cal G}_1$, 
that is, they are not $\theta$--independent, they are easily 
seen to simply mix into $\theta$--dependent linear combinations of themselves
as $\theta$ is varied, and so the constraints of Eq.~(45a) are 
$\theta$--independent.  Taking $\theta=0$ and making a change of 
variable $\vec x \to g^{-1} \cdot \vec x$, one
can explicitly solve the constraints to give a unique solution $g=g_I$,
$$\eqalign{
&g_{I11}=\left({\mu_{20} \over \mu_{00}} \right)^{1/2}~, ~~~~~g_{I12}=0~, \cr
&g_{I21}={\mu_{11} \over (\mu_{20} ~\mu_{00})^{1/2}}~, ~~~
g_{I22}=\left( {\mu_{02} ~\mu_{20}- \mu_{11}^2  \over \mu_{20} \mu_{00} } 
\right)^{1/2} ~. \cr
}\eqno(46)$$
Since the Schwartz inequality implies that $\mu_{11}^2 \leq \mu_{02} 
~\mu_{20}$,
the matrix element $g_{I22}$ is always a real number.  
[Equation (46) assumes that $\mu_{20}$ is nonzero; if $\mu_{20}$ vanishes and
if $I$ is not identically zero, then the moment $\mu_{02}$ will be 
nonvanishing, and
so one can apply Eq.~(46) after first rotating the image by 90 degrees.]  
The normalizing transformation for the subgroup $S_2$ is now constructed as 
$$\vec N_{2I}(\vec x)=g_I \cdot \vec x  ~. \eqno(47)$$
By a lengthy algebraic calculation, one can verify that under a general proper
(i.e., positive determinant) affine transformation $S$, the normalizing
matrix $g_I$ transforms as
$$g_I \to g_{I_S}= G^{-1} ~g_I ~R^{\prime} ~, \eqno (48a)$$
with $R^{\prime}$ a rotation matrix which is a 
complicated function of the
matrix elements of $g_I$ and of $G$.  Hence under the action of $S$, the
normalizing transformation for ${\cal G}_2$ becomes
$$\eqalign{
&\vec N_{2I_S}(\vec x)=g_{I_S} \cdot \vec x  \cr
&=(G^{-1} ~g_I ~R^{\prime}) \cdot \vec x = \vec S^{-1}(\vec N_{2I}
(R^{\prime} \cdot 
\vec x )) ~,  \cr
}\eqno(48b)$$ 
 in agreement with Eq.~(22c).  To obtain
an affine normalized image, one of course does not need the explicit form
of $R^{\prime}$; one first forms the partially normalized image
$$\tilde I(\vec x) = I(\vec N_{2I}(\vec x)) ~, \eqno(49a)$$
and then normalizes with respect to rotations as in Eq.~(36) of 
Subsec. 3C, to get the final normalized image
$$\hat I(\vec x)= I(\vec N_{2I}( \vec M_I(\vec x)_{\theta_{\tilde I}})) ~. 
\eqno(49b) $$
The map $\vec M_I$ is constructed only from affine invariants of the image; 
the
simplest choice is $\vec M_I(\vec x) = G_0 \cdot \vec x$, with $G_0$ a fixed 
affine transformation which can be chosen to give the affine normalized
version of one particular pattern a specified form.  With this choice of
$\vec M_I$, the normalization integral $\int ~d^2x \hat I(\vec x)$ required
for contrast normalization is explicitly given in terms of central moments by
$$\int ~d^2x \hat I(\vec x)={\mu_{00} \over \det G_0 \det g_I}
={\mu_{00}^2 \over \det G_0 ~~(\mu_{02} ~\mu_{20} - \mu_{11}^2)^{1/2} } ~.
\eqno(50) $$
\parindent=25pt
                                             
An alternative way to normalize the homogeneous affine transformations is
to use the subgroup factorization [c.f. Eq.~(5c)]
$$\eqalign{
&\vec S_2(\vec x)=g^{\prime} \cdot \vec x~, 
~~~~\vec S_1(\vec x)=\lambda ~\vec x_{\theta}, \cr
&\vec S(\vec x)=\vec S_2(\vec S_1(\vec x))= g^{\prime} \cdot 
\lambda~\vec x_{\theta}
=G \cdot \vec x~,  \cr
} \eqno(51a)$$
with $g^{\prime}$ restricted to have both zero upper right diagonal
matrix element and unit determinant.  Since $g^{\prime}$ and ${\cal G}_2$ 
now contain only
two parameters, and since ${\cal G}_1$ now includes both rotations and
scalings, partial normalization with respect to ${\cal G}_2$ requires two
constraints which must be both rotation and scaling invariant.  Inspecting
Eq.~(45b), we see that an obvious choice of constraint functionals is now
$$\eqalign{
&C_1^{\prime}[I_S]=\int ~d^2x ~(x_2^2 -~x_1^2) ~I(G \cdot \vec x)  ~, \cr
&C_2^{\prime}[I_S]=\int ~d^2x ~x_1 ~x_2 ~I(G \cdot \vec x)    ~. \cr
}\eqno(51b)$$
Again, although these functionals are not rotation and 
(in the case of  $C_2^{\prime}$) scale invariant, the constraints 
$C_1^{\prime}
=0,~C_2^{\prime}=0$ are invariant, and solving them gives the not surprising
result
$$g^{\prime}_I=(\det g_I)^{1/2} ~g_I ~,  \eqno(51c)$$
with $g_I$ as given in Eq.~(46).  The normalizing transformation for the
subgroup $S_2$ is now constructed as in Eq.~(47), with $g_I^{\prime}$
replacing $g_I$, and the partially normalized image is again given 
by Eq.~(49a),
but now the final step leading to a fully affine normalized image 
consists of a further combined normalization with respect to
rotation and scaling of the type described in Subsec.~3E.
\parindent=0pt
\bigskip
\centerline{\bf 4.~~Viewing Transformations With 
Numerically Solvable Constraints}
\parindent=25pt

In this section we continue with the application of the general 
image normalization 
methods of Sec.~2 to the
viewing transformations of a planar object, focusing on cases in    
which the constraints are not all algebraically solvable, so that 
iterative numerical methods are needed.   We then go on to consider some 
other normalization problems of interest, that can also be solved by   
iterative methods.  

\parindent=0pt
A.  {\it Projective transformations.}  So far we have discussed linear
transformations $\vec S(\vec x)$, which within the general normalization 
framework of Sec. 2 lead to algebraically solvable constraints.   
We turn now to nonlinear transformations,
to which the general analysis also applies, beginning with the 
planar projective transformation for which $\vec S(\vec x)$ is given by
$$\vec S(\vec x)= {\sum_{m=1}^2 ~G_{nm} ~x_m + t_n \over
1+\sum_{m=1}^2 ~p_m ~x_m}~. \eqno(52a)$$
We again use the subgroup decomposition method, writing 
$$\vec S(\vec x) =\vec S_2(\vec S_1(\vec x))~, \eqno(52b)$$
with $S_2 \in {\cal G}_2$ a restricted projective transformation 
and $S_1 \in {\cal G}_1$ an affine 
transformation, as in Eq.~(3) and Eq.~(2) respectively.  
Since ${\cal G}_2$ is a two--parameter Lie
group, we need two constraints, which must be invariant under the action
of the affine transformations of ${\cal G}_1$, to partially normalize 
the image.
We have not been able to find two simple 
constraint functionals which yield algebraically solvable 
affine invariant constraints
when equated to zero, which would be the analog of our previous 
two applications
of the subgroup method.  Instead, we work with  
constraint functionals which are
fully affine and contrast invariant, as obtained by the 
algebraic methods of Hu [11] and Reiss [14], which because of their 
complexity must be solved  
numerically.  (Alternatively, one could formulate the projective constraints 
using two independent affine invariants constructed by the affine 
normalization procedure of the preceding section, 
again solving the constraints numerically.  
We emphasize that in either case, the constraints 
used for projective normalization are {\it not} projective invariants, but 
only invariants under the much simpler affine subgroup of the full projective 
group.)  Using only third and lower central moments,
one can form the following three functionals of the image which are affine
and contrast invariant, and which are non--singular (in fact vanishing) for
images with both $x_1$ and $x_2$ reflection symmetry,
$$\eqalign{
&\Psi_1[I]={\mu_{00}^2 ~I_2 \over I_1^3}~,~~~\Psi_2[I]={\mu_{00} ~I_3 \over
I_1^2}~,~~~\Psi_3[I]={\mu_{00} ~I_4 \over I_1^3}~, \cr
&I_1=AC-B^2~, \cr
&I_2=(ad-bc)^2-4(ac-b^2)(bd-c^2)~, \cr
&I_3=A(bd-c^2)-B(ad-bc)+C(ac-b^2)~, \cr
&I_4=a^2C^3-6abBC^2+6acC(2B^2-AC)+ad(6ABC-8B^3)+9b^2AC^2\cr
&~~~~~-18bcABC+6bdA(2B^2-AC)+9c^2A^2C-6cdA^2B+d^2A^3 ~, \cr
&A=\mu_{20}~,~~~B=\mu_{11}~,~~~C=\mu_{02}~, \cr
&a=\mu_{30}~,~~~b=\mu_{21}~,~~~c=\mu_{12}~,~~~d=\mu_{03}~.  \cr
} \eqno(53a)$$
For example, $C_1[I]=C_2[I]=0$ could be used as constraints, with
$$C_1[I]=\Psi_1[I] - \Psi_1^0~, ~~~ C_2[I]=\Psi_2[I] 
- - \Psi_2^0~,
\eqno(53b)$$
{\it provided} the numerical target values $\Psi_1^0, \Psi_2^0$ fall within
the ranges taken by $\Psi_1, \Psi_2$ for the image $I$ being normalized.
Since the restricted projective transformation 
$$\vec S_2(\vec x) = {\vec x \over 1+ \vec p \cdot \vec x}  \eqno(54a)$$
depends nonlinearly on the parameter $\vec p$, we cannot algebraically 
solve the constraints to find the normalizing parameter $\vec p_I$, but this
can be readily done numerically by an  iterative method.
The partial normalization transformation and partially normalized image are
now given by
$$\eqalign{
&\vec N_{2I}(\vec x)= {\vec x \over 1+ \vec p_I \cdot \vec x}~,  \cr
&\tilde I(\vec x)=I(\vec N_{2I}(\vec x))~.  \cr
}\eqno(54b)$$
Finally, one must do a further affine normalization, as in Subsecs. 3B and 3F, 

to get an image normalized with respect to 
the full planar projection group.  If the
initial image is not well--centered on the raster, it may be advantageous to
also do an affine normalization {\it before} the restricted projective partial
normalization; this does not affect the results provided a second affine
normalization is still done as the final normalization step.
\parindent=25pt

The fact that one must know the range of $\Psi_1, \Psi_2$ to pick target
numerical values for normalization may prove a significant limitation, since
it is likely that there is no universal pair of target values which is
guaranteed to be attainable for any image.  (For a discussion of related 
problems with projective normalization, see \.Astr\"om [1].)  
Consequently, it may be necessary to have a preliminary classification of the 
viewed
object before attempting projective normalization.  However, this may not
be a problem in some applications, as for example when an approaching object
is tracked and can be classified when it is still far enough away for the
affine approximation to the general projective transformation to be accurate.
As the object gets closer, knowledge of its class can be used to determine
the constraints to be used in projective normalization, and the values of
$\vec p_I$ and the affine parameters obtained from projective normalization 
can then be used to
deduce information about the object's absolute orientation.   Another option, 
in
applications where  preliminary classification by an affine normalizing
classifier is feasible, 
is to use optimization of the match $M$ through 
the classifier to supply the projective constraints;
that is, the constraints are taken as
$${\partial M \over \partial p_1}={\partial M \over \partial p_2}=0~,
 \eqno(55)$$
which are solved by iteration on the restricted projective parameter $\vec p$
to determine $\vec p_I$.
\parindent=0pt

B. {\it Similarity and 
Affine Normalization of Partially Occluded Planar Curves.}  
So far we have discussed only the normalization of non-occluded images, 
but the general methods formulated in Sec. 2 have been extended by Adler 
and Krishnan [3] to the 
more realistic problem [4], [5] of the similarity and affine normalization of 
a 
partially occluded planar curve, such as that characterizing the boundary 
of a partially occluded planar object.  Since full details and 
illustrative numerical results are given 
in [3], we give here only a sketch of the strategy.  Consider, for 
simplicity, the special case in which one has a curve segment, 
with an identifying 
point $P$, distorted by an affine transformation.  One can construct  an 
affine normalization by the method of Sec. 3, by forming constraints 
using second moments of the curve integrated along a finite segment from 
$P$ to some neighboring point $P^{\prime}$.  To specify this integration 
segment in an affine invariant way, reference [3]  
requires that the normalized image of the segment have some 
specified reparameterization invariant arc-length, 
giving one additional constraint that must be solved by numerical iteration.  
The resulting normalization procedure for partially occluded planar curves  
normalizes against affine transformations using as input only first and 
second parametric derivatives, i.e., only information about the tangent 
vector and the curvature of the curve.    This example shows how the     
general methods of this paper can be used as modules in iterative procedures   

to solve new, previously unsolved, classes of normalization problems.  

C.  {\it Flexible template normalization.}  The nonlinear projective image
transformations which we have just discussed are only one example of much 
more general nonlinear distortions which can make an observed image
differ in form from the standard prototype for its class.  Examples of such
distortions include non--planar geometric effects when a character to be
recognized is printed on a curved surface, variations among
hand lettered characters produced by different individuals, 
and variations in facial geometry as a result
of changes in facial expression.  An attractive proposal for dealing with such
distortions is the use of ``flexible'' or ``deformable'' templates 
(see, e.g. [9]), and the general normalization
methods of Sec. 2 give a possible means for their implementation.  We consider
briefly in this subsection the case of distortions which can be modeled as
an image transformation
$$I(\vec x) \to I(\vec T(\vec x)) ~,  \eqno(56a)$$
where $\vec x^{\,\prime} =\vec T(\vec x)$ is a general nonlinear remapping or
diffeomorphism of the image coordinate $\vec x$.  Such 
diffeomorphisms form a group,
and so the general analysis of Sec. 2 formally applies, but since the general
diffeomorphism group has an infinite number of parameters, this observation
is of little practical use without making further assumptions.  Let us now
suppose that the predominant nonlinear distortions can be treated as small
in magnitude, and are well represented by a few terms in an appropriate
complete expansion basis.  A concrete example would be nonlinear distortions
described by the transformation
$$\vec T(\vec x) = \vec t + G \cdot \vec x + H \cdot \vec x \vec x
+J \cdot \vec x \vec x \vec x ~, \eqno(56b) $$
with the $H$ term shorthand for the vector with components $\sum_{mp}
~H_{nmp} x_m x_p$, and similarly for the $J$ term.  When $H,J$ are effectively
of order unity, the transformations of Eq.~(56b) do not form a group, 
since iteration of
the transformation leads to fourth and higher order terms in $\vec x$.  
However,
if $H,J$ are small enough for terms quadratic and higher order in $H,J$ to be
neglected, the transformations of Eq.~(56b) do, within the first order
approximation, form a group, and the normalization methods of Sec. 2 become
applicable.  One could then proceed by first constructing an affine 
preprocessing
classifier by the methods of Subsecs. 3A--F, thus normalizing for the linear
transformation given by the first two terms of Eq.~(56b).   One would then 
normalize with respect to all of Eq.~(56b) by iterating  on the 
coefficients $H,J$ to try to get an optimal unique classification through
this classifier, using the cost function
$$C= ||{\rm classifier ~mismatch}|| + ||H|| + ||J|| ~, \eqno(56c)$$
with the terms in Eq.~(56c) giving respectively measures of the magnitude 
of the classifier 
mismatch for the class being considered, the magnitude
of the coefficients $H$, and the magnitude of the coefficients $J$.  Clearly,
a similar method could be applied to the expansions of $\vec T$ on any 
polynomial (and perhaps more general) basis, provided the 
truncated basis is left
invariant in form within a suitable first order approximation, for which the 
affine transformations form the zeroth order approximation.

D.  {\it Normalization of an Image on a Sphere.} As a final application of the
methods of Sec.~2, we briefly discuss the normalization of an image 
$I(\Omega)$
defined on the surface of a sphere of radius $R$ and angular variables 
$\Omega$, with respect to the group of rotations $S$ of the sphere. 
We shall confine ourselves to the simplest case, in which 
the image $I$ has no special symmetries which make the relevant constraint
integrals vanish.  The normalization can be carried out in two steps.  The
first step imposes the constraint
$$\vec C[I_S]={\int ~d\Omega ~I_S(\Omega) ~\hat n_{\Omega} \over
|\int ~d\Omega ~I_S(\Omega) ~\hat n_{\Omega}|} -(0,0,1) =0~, \eqno(57)$$
with $\hat n_{\Omega}$ the outward pointing three dimensional unit normal to 
the
sphere at $\Omega$.  This rotates the sphere so that the positive $x_3$ axis
(the north polar axis of the sphere) 
passes through the center of mass (calculated in spherical geometry) of the
image.  The second step consists of a rotational normalization with respect
to azimuth (or longitude) using the formulas of Subsec.~3C, in 
which dependences on $|\vec x|$ are replaced by 
dependences on the spherical polar angle (or latitude).  In the group 
contraction limit in which the sphere radius $R$ approaches infinity while the
dimension of the region of support of the image remains bounded, this 
normalization recipe reduces to that of Subsecs.~3A, C 
for combined translational
and rotational normalization of a planar image.
\parindent=25pt
\bigskip 
\centerline{\bf 5.~~ Summary and Discussion}  

We have given a general normalization method for viewing transformations 
of planar images, based on imposing a set of constraints equal in number 
to the parameters of the viewing transformation group, the solution of 
which gives the parameters of the normalizing transformation.  In 
Sec. 3 we discussed linear viewing transformations, for which algebraically  
solvable constraints can be given.  In Sec. 4 we discussed more complex 
situations, in which some of the constraints cannot be solved 
algebraically, but 
can be solved by numerical iterative methods.  
Although the normalization methods of Subsecs. 3A--F and 4A-B were all 
based on the use of moments or other weighted integrals of the image 
to construct the constraints, the general analysis of Sec. 2 does
not require this. 
Alternative methods include setting
the scale normalization by a determination of the outer boundary of the 
image (as in [10], [16]), and setting the rotational 
normalization angle after scale normalization by using the maximum
of $I$ in an annular ring of given radius, both of which are methods 
that use local image features instead of weighted integrals over the image.    
 
The most convenient set of constraints will, in practice, depend on the 
specifics of the invariance problem being analyzed.  In general, the larger 
the number of constraints that can be solved algebraically, and the 
smaller the number that require numerical solution, the more computationally 
efficient will be the resulting normalization method.  For this reason, 
we have given particular emphasis to subgroup methods, that express some 
of the constraints needed for more complex normalization problems 
in terms of those already constructed for simpler normalization problems, 
for which algebraic solution methods are available.  

In conclusion, we emphasize that according to the general theory
established in Sec. 2 and illustrated in Secs. 3 and 4, 
{\it any set of constraints that uniquely 
breaks the viewing transformation group invariance suffices to construct 
a normalization, and 
thereby to yield all viewing transformation invariants.}  The difference 
between normalizations constructed using alternative sets of constraints 
will always be representable by a residual mapping of the image, 
depending on the image only through viewing transformation invariants.

\bigskip
\centerline{\bf Acknowledgments}
I wish to thank J. Atick, E. Baum, F. Bedford, A. Bruckstein, 
I. Chakravarty, B. Dickenson, 
H. Freeman, R. Held, S. Kulkarni, S.Y. Kung, G. Miller, and N. Redlich for
helpful conversations.
This work was supported in part by the Department of Energy under
Grant \#DE--FG02--90ER40542.

\vfill\eject
\centerline{\bf References}
\bigskip
\noindent
[1] K. \.Astr\"om,  ``Fundamental Difficulties with Projective Normalization 
of Planar Curves,'' {\it Applications of Invariance in Computer Vision},  
J. L. Mundy, A. Zisserman, and D. Forsyth, Eds.  
Berlin: Springer--Verlag, 1993, pp. 199--214.
\bigskip
\noindent
[2] Y. S. Abu--Mostafa and D. Psaltis, ``Image Normalization by Complex 
Moments'', {\it IEEE Trans. Pat. Analy. and Mach. Intell.}, vol. 7, no. 1, 
pp. 46--55, Jan. 1985.
\bigskip
\noindent
[3] S. L. Adler and R. Krishnan, ``Similarity and Affine Normalization 
of Partially Occluded Planar Curves Using First and Second Derivatives'', 
{\it Pattern Recognition}, vol. 31, no. 10, pp. 1551-1556, 1998. 
\bigskip                    
\noindent 
[4] A. M. Bruckstein, R. J. Holt, A. N. Netravali, and T. J. Richardson, 
``Invariant Signatures for Planar Shape Recognition under 
Partial Occlusion,'' {\it CVGIP: Image Understanding},  vol. 58, no. 1, 
pp. 49--65, July 1993.
\bigskip 
\noindent
[5]  A. M. Bruckstein and A. N. Netravali, ``On differential invariants 
of planar curves and recognizing partially occluded planar shapes,'' 
{\it Ann. Math. and Art. Intell.}, vol. 13, pp. 227--250, 1995.
\bigskip
\noindent
[6] H. Dirilten and T. G. Newman, ``Pattern Matching Under Affine 
Transformations,''{\it IEEE Trans. Comp.}, 
vol. 26, no. 3, pp. 314--317,  March 1977.
\bigskip
\noindent
[7] M. Ferraro, ``Invariant Pattern Representations and Lie Groups Theory,''
{\it Image Mathematics and Image Processing, Advances in Electronics and 
Electron Physics}, vol. 84,  P. W. Hawkes, Ed. Boston: Academic
Press, 1992, pp. 131-195.
\bigskip
\noindent
[8] R. Gilmore, {\it Lie Groups, Lie Algebras and Some of 
Their Applications}.
New York:  Wiley, 1974.  Reprinted by Malabar, Florida: Krieger Pub. Co., 
1994.  
\bigskip
\noindent
[9] U. Grenander, {\it General Pattern Theory}. Oxford: Clarendon Press, 1993.
\bigskip
\noindent
[10] S. L. Horowitz,  ``Method and Apparatus for Generating Size and 
Orientation Invariant Features,'' {\it U.S. Patent no.} 4,989,257, Jan. 1991.
\bigskip
\noindent
[11] M. K. Hu, ``Visual Pattern Recognition by Moment Invariants,'' 
{\it IRE Trans. Inf. Th.}, vol. 8, no. 2 , pp. 179--187, Feb. 1962.
\bigskip
\noindent
[12] R. Lenz, ``Group Theoretic Methods in Image Processing,'' 
{\it Lecture Notes in Computer Science}, no. 413,  G. Goos and J. Hartmanis,  
Eds. Berlin: Springer--Verlag, 1987, pp. 1--139.
\bigskip
\noindent
[13]  D. H. Sattinger and O. L. Weaver, {\it Lie Groups and Algebras 
with Applications to Physics, Geometry, and Mechanics}.  New York: 
Springer, 1986.  
\bigskip
\noindent
[14] T. H. Reiss, ``The Revised Theorem of Moment Invariants,'' 
{\it IEEE Trans. Pat. Anal. and Mach. Intell.}, vol. 13, no. 8, pp. 830--834, 
Aug. 1991.
\bigskip
\noindent
[15] T. H. Reiss, {\it Recognizing Planar Objects Using Invariant Features}. 
Berlin: Springer--Verlag, 1993.
\bigskip
\noindent
[16] Y. Y. Tang, H. D. Cheng, and C. Y. Suen, 
``Transformation--Ring--Projection (TRP) Algorithm and its VLSI 
Implementation,''                             
{\it Character \& Handwriting Recognition}, P. S. P. Wang, Ed.  
Singapore:  World Scientific, 1991, pp. 25-56.
\vfill
\eject
\bigskip
\bye